# Stochastic Logic Programs: Sampling, Inference and Applications

**James Cussens**
Department of Computer Science, University of York
Heslington, York, YO10 5DD, UK
jc@cs.york.ac.uk

## Abstract

Algorithms for exact and approximate inference in stochastic logic programs (SLPs) are presented, based respectively, on variable elimination and importance sampling. We then show how SLPs can be used to represent prior distributions for machine learning, using (i) logic programs and (ii) Bayes net structures as examples. Drawing on existing work in statistics, we apply the Metropolis-Hasting algorithm to construct a Markov chain which samples from the posterior distribution. A Prolog implementation for this is described. We also discuss the possibility of constructing explicit representations of the posterior.

## 1 Introduction

A stochastic logic program (SLP) is a probabilistic extension of a normal logic program that has been proposed as a flexible way of representing complex probabilistic knowledge; generalising, for example, Hidden Markov Models, Stochastic Context-Free Grammars and Markov nets (Muggleton, 1996; Cussens, 1999). However, we need to ask (i) whether this increase in flexibility is needed for any real problems and (ii) whether reasonable algorithms exist for inference and learning in SLPs.

In this paper we give a number of approaches to approximate and exact inference in SLPs, focusing mostly on sampling. More importantly, we apply SLPs to an important problem—Bayesian machine learning—which would be difficult to handle with simpler representations.

The paper is organised as follows. Section 2 contains essential definitions from logic programming. Section 3 shows we can use SLPs to define three different sorts of distributions, but focuses on the appealingly simple *loglinear model*. Sections 4 and 5 give two quite different ways of improving sampling for the loglinear model, being based on Prolog and importance sampling, respectively. Section 6 briefly shows how variable elimination can be used for exact inference in SLPs. After showing how to extend SLPs with non-equational constraints in Section 7, we can finally bring much of the preceding work together in Sections 8 and 9 to show how SLPs can be used to represent distributions over complex model spaces, such as is required for 'really Bayesian' machine learning. Section 10 contains conclusions and pointers to possible future work.

## 2 Logic Programming Essentials

An overview of logic programming can be found in (Cussens, 1999). Here, lack of space means that only the most important definitions are flagged. An *SLD-derivation* of a goal $G_0$ using a logic program $\mathcal{P}$ is a (finite or infinite) sequence $\{\text{State}_j\}_j$, each element of which is either a 4-tuple $(G_j, A_j, C_j, \theta_j)$ or the empty goal $\Box$ where

- $A_j$ is the selected atom of goal $G_j$;

- $C_j$ is the selected input clause in $\mathcal{P}$, with its variables renamed so that $C_j$ and $G_j$ have no variables in common;

- $\theta_j$ is the most general unifier of $A_j$ and $C_j^+$ (the head of $C_j$) or fail if they can not be unified.

- If $\theta_j = $ fail then $\text{State}_{j+1} = $ fail. Otherwise, $G_{j+1}$ is the result of replacing $A_j$ by $C_j^-$ (the body of $C_j$) in $G_j$ and then applying $\theta_j$ to the result. If $G_{j+1} = \Box$ then $\text{State}_{j+1} = \Box$.

An *SLD-refutation* is a finite SLD-derivation ending in the empty goal $\Box$. The *SLD-tree* for a goal $G$ is a tree of goals, with $G$ as root node, and such that the children of any goal $G'$ are goals produced by one resolution step using $G'$ ($\Box$ and fail have no children). Fig 6 shows an SLD-tree. A *computed answer* for a goal is a substitution for the variables in $G$ produced by an SLD-refutation of $G$.

## 3 Defining Distributions with SLPs

A stochastic logic program (SLP) is a logic program where some of the predicates (the *distribution-defining* or *probabilistic predicates*) have non-negative numbers attached to the clauses which make up their definitions. We denote the *label* for a clause $C_i$ by $l_i$. We also write $\lambda_i = \log l_i$, and we denote the parameter vector containing all the $\lambda_i$ by $\lambda$. In this paper we will consider only *normalised* SLPs where labels for the clauses making up the definition of a predicate sum to one.

The basic idea is that clause labels probabilistically influence which input clause is chosen as a derivation proceeds, thus defining a distribution over derivations from which distributions over (i) refutations and (ii) variable bindings may be derived. We begin by restricting attention to *pure* SLPs, where all predicates have labelled definitions, postponing the impure case until Section 7.

### 3.1 Loglinear Model

Given an SLP $\mathcal{S}$ with parameters $\lambda$ and a goal $G_0$ we can sample SLD-derivations for $G_0$ using *loglinear sampling* as follows:

> *Loglinear sampling*: Any computation rule may be used to select the atom $A_j$ from the current goal $G_j$. The next input clause $C_j$ is chosen with probability $l_j$ from those clauses in $\mathcal{S}$ with the same predicate symbol in the head as $A_j$. We stop when we produce either `fail` or $\square$.

Let $R(G)$ denote the set of refutations of the goal $G$ (i.e. derivations that end in $\square$). Let $\nu_i(r)$ be the number of times labelled clause $C_i$ is used in refutation $r$. The loglinear distribution $p_{(\lambda,\mathcal{S},G)}$ over refutations $r$ of the goal $G$ is

$$\forall r \in R(G) : p_{(\lambda,\mathcal{S},G)}(r) = \frac{\psi_{(\lambda,\mathcal{S},G)}(r)}{Z_{(\lambda,\mathcal{S},G)}}$$

where

$$\psi_{(\lambda,\mathcal{S},G)}(r) = \prod_j l_j = \prod_i l_i^{\nu_i(r)}$$

$$Z_{(\lambda,\mathcal{S},G)} = \sum_{r \in R(G)} \psi_{(\lambda,\mathcal{S},G)}(r)$$

The loglinear distribution over all derivations, which includes infinite ones and ones ending in `fail`, is similar except that $Z_{(\lambda,\mathcal{S},G)}$ is guaranteed to equal one.

Computing the unnormalised probability (*potential*) $\psi$ of an already given refutation is efficient, since we just multiply clause labels as we go. Loglinear sampling to produce a refutation is less efficient. It is essentially the same as *forward sampling* in Bayesian networks and suffers from the same defect: inefficiency when sampling from a distribution conditional on evidence. In the case of SLPs the same problem arises when, as is usually the case, we are only interested in refutations—which are derivations conditional on their last goal being the empty goal. Sampling from the full set of derivations, including failed ones, is easy.

The problem is the restricted information that is readily available when selecting the next input clause. Let us call $Z_{(\lambda,\mathcal{S},G)}$ the *weight* of the SLD-tree under $G$ or briefly the weight of $G$. If we had the weights of the potential successor goals each possible input clause would give, then we could sample easily from the loglinear model by simply making choices in proportion to these weights. The stochastic search rule thus implemented would never follow a failure branch, since such branches have zero weight.

If $|R(G)|$ is reasonably small then we can simply use Prolog to find all $r \in R(G)$ together with their probabilities $p_{(\lambda,\mathcal{S},G)}(r)$, store this information in a simple table and then sample the $r \in R(G)$ according to $p_{(\lambda,\mathcal{S},G)}(r)$. In some applications the central problem is searching a possibly very large SLD-tree for a few refutations—for this we can use normal logic programming approaches. In particular, chart-based and/or bottom-up approaches to finding refutations will often be appropriate. In statistical computational linguistics, refutations amount to parses, and there is a substantial body of work on finding parses and their probabilities. Usually interest is confined to finding the single most likely parse. For example, (Riezler, 1998) presents an approach to this problem that trades efficiency for optimality in a generalisation of the Viterbi algorithm using Earley deduction. (Muggleton, 2000) presents an algorithm that enumerates refutations in order of decreasing probability. (Muggleton, 2000) also discusses how to *unfold* an SLP to make it 'more tabular'.

However, many interesting applications will involve goals with large, even infinitely many refutations. In such cases, both loglinear sampling and "all refutations sampling" can both be very inefficient. In the next two sections, we consider alternatives which lead to more efficient sampling.

### 3.2 Unification-constrained Model

*Unification-constrained sampling* finds refutations more efficiently than loglinear sampling because it only chooses clauses from amongst those that unify with the selected atom. It effects a one-step lookahead by examining potential input clauses before selecting one. Let $\text{unif}(A_j)$ be the set of clauses whose heads unify with the atom $A_j$.

> *Unification-constrained sampling*: The selected atom $A_j$ is always the leftmost atom in the current goal $G_j$. The next input clause $C_j$ is chosen from $\text{unif}(A_j)$ with probability proportional to $l_j$. We stop when we produce either `fail` or $\square$.

This is the original sampling mechanism defined by (Muggleton, 1996). It is a theorem of logic programming that normal SLD-refutation is essentially unaltered by changing the atom selection rule (although drastic changes in efficiency can occur). If we delay selecting an atom $A$, all that changes is that $A$ may be more instantiated when we eventually do select it. Unfortunately, this means that delaying the selection of $A$ will alter $\text{unif}(A)$ and so we are obliged to fix the selection rule in unification-constrained sampling so that a single distribution is defined.

The unification-constrained distribution $p^u_{(\lambda,S,G)}$ over refutations for a goal $G$ is

$$\forall r \in R(G) : p^u_{(\lambda,S,G)} = \frac{\psi^u_{(\lambda,S,G)}(r)}{Z^u_{(\lambda,S,G)}}$$

where

$$\psi^u_{(\lambda,S,G)}(r) = \prod_j \frac{l_j}{\sum_{C_{j'} \in \text{unif}(A_j)} l_{j'}}$$

$$Z^u_{(\lambda,S,G)} = \sum_{r \in R(G)} \psi^u_{(\lambda,S,G)}(r)$$

Computing the unnormalised probability (*potential*) $\psi_u$ of a refutation is only slightly less easy than with the loglinear model, since all the values $l_{j'}$, for $C_{j'} \in \text{unif}(A_j)$, can be quickly found as the refutation proceeds.

### 3.3 Backtrackable Model

Unification-constrained sampling stops as soon as it reaches `fail`, *backtrackable sampling* backtracks upon failure, and so will generally be a more efficient way of sampling refutations. Backtrackable sampling is essentially Prolog with a probabilistic clause selector.

> *Backtrackable sampling*: The selected atom $A_j$ is always the leftmost atom in the current goal $G_j$. The next input clause $C_j$ is chosen with probability $l_j$. If we produce `fail` then we backtrack to the most recent choice-point, delete the choice of input clause that led to failure and choose from amongst the surviving choices with probability proportional to clause labels. If no choices remain then we backtrack further until a choice can be made or return `fail`. We stop if we produce □ or `fail`.

Let $\text{succ}(A_j)$ be the set of input clauses which lead to at least one refutation. The backtrackable distribution $p^b_{(\lambda,S,G)}$ over refutations for a goal $G$ is

$$\forall r \in R(G) : p^b_{(\lambda,S,G)}(r) = \frac{\psi^b_{(\lambda,S,G)}(r)}{Z^b_{(\lambda,S,G)}}$$

where

$$\psi^b_{(\lambda,S,G)}(r) = \prod_j \frac{l_j}{\sum_{C_{j'} \in \text{succ}(A_j)} l_{j'}}$$

$$Z^b_{(\lambda,S,G)} = \sum_{r \in R(G)} \psi^b_{(\lambda,S,G)}(r)$$

Computing the unnormalised probability (*potential*) $\psi_b$ of a refutation is, in general, hard; since we may have to explore a very large SLD-tree rooted at $G_j$ to identify the set $\text{succ}(A_j)$.

Comparing the loglinear, unification-constrained and backtrackable models we see there is a trade-off between ease of sampling a refutation, and ease of computing a potential for a given refutation. If only sampling is required and we are happy that the order of literals in clauses matters, then the backtrackable model makes sense. However, the loglinear model has attractively simple mathematical properties which we will exploit in the MCMC application. Fortunately, loglinear sampling can be sped up as the next section shows.

## 4 Improving Loglinear Sampling

(Muggleton, 1996) explicitly introduced SLPs as generalisations of Hidden Markov models (HMMs) and Stochastic Context-Free Grammars (SCFGs). Comparing SLPs to HMMs we see that in SLPs: (i) the states of an HMM are replaced by goals (ii) the outputs of an HMM are replaced by substitutions (iii) concatenation of outputs is replaced by composition of substitutions (iv) outputs (substitutions) are generated deterministically and (v) state transition probabilities are given by clause labels. It is also more natural in SLPs to associate outputs (substitutions) with transitions between states (goals) rather than with states themselves.

The connection between SLPs and SCFGs is even closer. Consider $S_1$, the SCFG in Fig 1 which has been implemented as an SLP. We can generate sentences using loglinear sampling with the goal :- s(A,[]). Since $S_1$ is an SCFG, the query will always succeed, even though we do not allow backtracking in the loglinear model. Suppose

```
1:s(A,B)  :- n(A,C), v(C,D), n(D,B).
0.4:n([joe|T],T).   0.6:n([kim|T],T).
0.3:v([sees|T],T).  0.7:v([likes|T],T).
```

Figure 1: $S_1$: An SCFG

now, that we are interested only in reflexive sentences. We then apply a *constraint* to the SCFG: replacing
```
1:s(A,B)  :- n(A,C), v(C,D), n(D,B).
```
with
```
1:s(A,B)  :- n(A,C), v(C,D), n(D,B)
             A=[N|T1], D=[N|T2].
```

or more concisely:
```
1:s([N|T1],B) :- n([N|T1],C),
    v(C,[N|T2]), n([N|T2],B).
```
Now we can not guarantee that `:- s(A,[]).` will always succeed, the grammar is no longer context-free. This means that sampling sentences from the new conditional distribution (conditional on the constraint being satisfied) is less efficient. We have to throw away derivations that are inconsistent with the constraint, just as with forward sampling in Bayes nets in the presence of evidence.

In the loglinear model we may select any atom from the current goal, which means that the order of literals in the bodies of clauses does not affect the distribution. However, since we will use Prolog to implement SLPs, we can exploit Prolog's leftmost atom selection rule to force constraints to be effected as early as possible. We do this by simply moving the constraints leftwards so that Prolog encounters them earlier. This has the effect of producing fail as soon as our choice of input clauses has ensured that a derivation can not succeed. Fig 2 has an ordering of body literals for s/2 that ensures that a derivation fails as soon as we pick a second noun which is not the same as the first—we don't waste time choosing the verb. The moral is: *it is better to fail sooner than later.*

```
1:s([N|T1],B) :-
    n([N|T1],C),n([N|T2],B),v(C,[N|T2]).
0.4:n([joe|T],T).     0.6:n([kim|T],T).
0.3:v([sees|T],T).    0.7:v([likes|T],T).
```

Figure 2: $S_2$: A simple grammar

## 5 Importance Sampling for SLPs

SLPs are only required for complex distributions, where it is optimistic to depend on exact inference. Approximate inference can be based on sampling, where e.g. to estimate the probability of some event, we sample from the SLP and obtain the event's relative frequency. Unfortunately, even with the Prolog-based speed up given above, pure loglinear sampling can still be slow. However, since $p^u_{(\lambda,S,G)}$ is easier to sample from than the loglinear distribution $p_{(\lambda,S,G)}$, the obvious solution is to use *importance sampling* (Gelman et al., 1995). We produce samples from $p^u_{(\lambda,S,G)}$ and then weight them with the importance weights $p_{(\lambda,S,G)}(r)/p^u_{(\lambda,S,G)}(r)$ We have:

$$\frac{p_{(\lambda,S,G)}(r)}{p^u_{(\lambda,S,G)}(r)} = \frac{[Z_{(\lambda,S,G)}]^{-1} \prod_j l_j}{[Z^u_{(\lambda,S,G)}]^{-1} \prod_j \frac{l_j}{\sum_{C_{j'} \in \text{unif}(A_j)} l_{j'}}}$$

$$= \frac{Z^u_{(\lambda,S,G)}}{Z_{(\lambda,S,G)}} \prod \sum_{C_{j'} \in \text{unif}(A_j)} l_{j'}$$

We can update $\prod \sum_{C_{j'} \in \text{unif}(A_j)} l_{j'}$ as we go, so therefore it is easy to compute weighted samples for a particular goal, where the weights are known up to a normalising constant. For approximate inference, the unknown normalising constant will often cancel out. For example, it is frequently enough to estimate the ratio between probabilities.

## 6 Exact Inference in SLPs

Each refutation of a goal $G$ produces a computed answer—variable bindings for the variables in $G$. We can define a distribution over the computed answers for $G$ by marginalisation—we sum over all refutations that produce a computed answer. It is convenient to represent computed answers by atoms: Define the *yield* $Y(r)$ of a refutation $r$ of a unit goal $G =\leftarrow A$ to be $A\theta$ where $\theta$ is the computed answer for $G$ using $r$. Let $\{X/x, Y/y, W/f(V)\}$ be a computed answer for the goal $\leftarrow r(X,Y,W)$, then the corresponding yield is $r(x,y,f(V))$ and:

$$p_{(\lambda,S,\leftarrow r(X,Y,W))}(r(x,y,f(V))) = \frac{Z_{(\lambda,S,\leftarrow r(x,y,f(V)))}}{Z_{(\lambda,S,\leftarrow r(X,Y,W))}}$$

Note that computed answers need not be ground (in contrast to previous work) and that we have overloaded $p_{(\lambda,S,G)}$ so that now it denotes a distribution over yields as well as refutations. To do exact inference for the distribution over computed answers we need to compute ratios of goal weights (the $Z$ values). I have yet to find a way of computing ratios without computing the weights themselves, so here is a method for calculation of weights.

In computing $Z_{(\lambda,S,G)} = \sum_{r \in R(G)} \psi_{(\lambda,S,G)}(r)$, for some goal $G$, one could find all potentials, one by one and then add them up. However, in general, it is far more efficient to find the weights of *subgoals* and multiply these weights as we go. This is simply an incarnation of the well-known variable elimination algorithm, so we will just give a brief overview of how it applies to SLPs. The basic operation is to sum out the choice of input clause for selected atom $A_s$:

$$Z_{(\lambda,S,\leftarrow A_1,...,A_s,...,A_m)}$$
$$= \sum_{C \in \text{unif}(A_s)} l(C) Z_{(\lambda,S,(\leftarrow A_1,...,C^-,...,A_m)\theta)}$$

Here $\theta$ is the mgu of $A_s$ and $C^+$. By defining $Z_{(\lambda,S,\square)} = 1$, we can now compute goal weights more efficiently. However, if several of the $C \in \text{unif}(A_s)$ produce the same summand we will end up computing the same weight several times, once for each $l(C)$. For example, we compute $Z_{(\lambda,S,\leftarrow q(Y))}$ twice, when computing $Z_{(\lambda,S,\leftarrow p(X),q(Y))}$ in the following SLP:

```
0.6:p(a). 0.4:p(b). 0.3:q(a). 0.7:q(b).
```

We can avoid this sort of problem by decomposing goals as follows. If a goal $G = (G_1, G_2)$ is a conjunction of two subgoals $G_1$ and $G_2$ which do not share variables then

$$Z_{(\lambda,S,G)} = Z_{(\lambda,S,(G_1,G_2))} = Z_{(\lambda,S,G_1)} Z_{(\lambda,S,G_2)}$$

For goals, such as $\leftarrow p(X,Y), q(Y,Z)$, which can not be decomposed into subgoals without common variables, we are forced to find *splitting substitutions*. A substitution $\theta$ *splits* two goals $G_1$ and $G_2$ if $G_1\theta$ and $G_2\theta$ do not share variables. Let $\Theta(G_1, G_2)$ be a set of splitting substitutions for $G_1$ and $G_2$ which includes all computed answers for the goal $(G_1, G_2)$ restricted to the common variables of $G_1$ and $G_2$. Then:

$$Z_{(\lambda,S,(G_1,G_2))} = \sum_{\theta \in \Theta(G_1,G_2)} Z_{(\lambda,S,G_1\theta)} Z_{(\lambda,S,G_2\theta)}$$

We would like to find a small $\Theta(G_1, G_2)$ fairly quickly. If each variable can only take a fairly small number of discrete values as is often the case in Bayesian nets, we can just go through each of these values. We end this section by noting that these computations need to be vectorised to return a (finite) distribution over bindings for a variable, rather than a single probability.

## 7  Impure SLPs

We can extend the definition of SLPs by going beyond conjunctions of equational constraints such as in $S_2$. Fig 3 shows $S_3$, an SLP for a fragment of French, where there is a constraint that adjectives and nouns agree on gender. The g/2 predicate that defines the constraint is unlabelled since it plays no part in defining the distribution except to cut out derivations which are inconsistent with it. When an unlabelled predicate is encountered in a derivation we only consider the first variable binding it produces (if any). Backtracking may be used to produce this first binding, but we may not return to seek another binding if we hit failure later on.

Equational constraints can be placed as early as we like. For other constraints, placing them too early can prevent permissible derivations from being found. The 2nd (commented out) version of s/2 in Fig 3 has the g(A, G) too early. Since backtracking is banned, only the first *il* binding will be found, overly constraining the value of A so e.g. *elle sera vieille* will never be produced. In the correct version we probabilistically (partially) instantiate the variable A, by our choice of input clause and only then effect the constraint on G. Since we are allowed to backtrack within the call to g/2 this call will always succeed. The key is that it must be the probabilistic predicates that choose the variable bindings that matter.

Effecting *negated constraints* too early can let through more derivations than is safe. The third version of s/2 starts with a negated goal that will succeed and produce no variable bindings—so no constraint is effected. Had this double negation been at the end of the clause it would have effected the desired constraint.

```
%Constraint too late - inefficient
%1:s(A,B) :- n(A,C), v(C,D), a(D,B),
%     g(A,G), g(D,G).

%Constraint too early - overconstrained
%1:s(A,B)  :- g(A,G), n(A,C),
%     a(D,B), g(D,G), v(C,D).

%Constraint too early - underconstrained
%1:s(A,B):- \+\+(g(A,G),g(D,G)),
%       n(A,C), v(C,D), a(D,B).

%Constraint  at the right time.
1:s(A,B) :- n(A,C), g(A,G),
    a(D,B), g(D,G), v(C,D).

0.4:n([il|T],T).      0.6:n([elle|T],T).
0.3:v([est|T],T).     0.7:v([sera|T],T).
0.2:a([vieux|T],T).   0.8:a([vieille|T],T).
g([il|_],m).     g([elle|_],f).
g([vieux|_],m).  g([vieille|_],f).
```

Figure 3: $S_3$: Gender agreement constraint

## 8  Machine Learning for Dogmatic Bayesians

> Finally, never forget that the goal of Bayesian computation is not the posterior mode, not the posterior mean, but a representation of the entire *distribution*, or summaries of that distribution such as 95% intervals for estimands of interest (Gelman et al., 1995, p.301) (italics in the original)

'Bayesian' approaches in machine learning do not live up to this exacting demand to represent the entire posterior, usually settling for just the posterior mode (MAP algorithms) or particular expectations (Bayes optimal classification). In this paper, we show how SLPs can be used to define priors representing a wide range of biases and constraints and also show how to sample from posteriors. Although we fall short of constructing (usable) explicit representations of the posterior, such a possibility can not be ruled out.

Our approach is based on the process prior approach for decision trees developed independently by (Chipman et al., 1998) and (Denison et al., 1998). Our presentation will follow that given by (Chipman et al., 1998). In short:

> Instead of specifying a closed-form expression for the tree prior, $p(T)$, we specify $p(T)$ implicitly by a tree-generating stochastic process. Each realization of such a process can simply be considered a random draw from this prior. Furthermore, many specifications allow for straightforward evaluation of $p(T)$ for any $T$ and can be effectively coupled with efficient Metropolis-Hastings search algorithms ... (Denison et al., 1998)

We can use Metropolis-Hastings to sample from the posterior distribution over trees, by choosing an initial tree $T^0$ and producing new trees as follows (where $p(Y|X,T)$ is just the likelihood with tree $T$) (Denison et al., 1998):

1. Generate a candidate value $T^*$ with probability distribution $q(T^i, T^*)$.

2. Set $T^{i+1} = T^*$ with probability
$$\alpha(T^i, T^*) = \min\left\{\frac{q(T^*, T^i)}{q(T^i, T^*)} \frac{p(Y|X, T^*)p(T^*)}{p(Y|X, T^i)p(T^i)}, 1\right\}$$
else set $T^{i+1} = T^i$

Because SLPs define distributions over *first-order atomic formulae*—the yields of refutations—they can easily represent distributions over model structures such as decision trees, Bayesian nets and logic programs. We will denote models using $M$, possibly superscripted. The simplest, but possibly very inefficient approach to defining priors over the model space with SLPs is as follows:
`model(M) :- gen(M), ok(M).`
`gen/1` generates possible models just like a SCFG generates sentences: there are no constraints so we never hit `fail`. `ok/1` is then a constraint which filters out models which we do not wish to include in the model space.

If `ok/1` rejects many of the models generated by `gen/1`, then it will be inefficient to sample from the prior and this inefficiency translates to inefficiency when running the Metropolis-Hastings algorithm. The solution, as explained in Section 4 is to move constraints as early as possible without altering the distribution. This has been done in the SLPs $S_4$ and $S_5$ of which fragments can be found in Fig 4 and Fig 5. These define priors over logic programs and Bayesian networks respectively. In each case, we simply define what a model is, using first-order logic, and then add labels to define a distribution over this model space. In $S_4$ we have constraints that we do not want an empty theory and that any generated rules have not previously been generated (`newrule/2`). We also have a 'utility' constraint `make_nice/2` which always succeeds and just rewrites the generated logic program into a more convenient form. In $S_5$ we assume that the variable `RVs` is always instantiated to a list of names of random variables, so that $S_5$ is used to define distributions of the form $p_{(\lambda, S_5, \leftarrow net([a,s,t], Net))}$, where all the probabilistic information is associated with `parents/3`. A more efficient version would push the `acyclic/1` constraint earlier.

```
model(LP) :- theory([],LP).
0.1:theory(Done,NicelyDone) :-
   \+ Done=[],
   make_nice(Done,NicelyDone).
0.9:theory(RulesSoFar,Done) :-
   rule(Rule),
   newrule(RulesSoFar,Rule),
   theory([Rule|RulesSoFar],Done).
```

Figure 4: Fragment of $S_4$, an SLP defining a prior over logic programs

```
model(RVs,Net) :- net(RVs,RVs,Net),
   acyclic(Net).
net([],_,[]).
net([H|T],RVs,Net) :-
   parents(H,RVs,Ps),
   append(Ps,TNet,Net),
   net(T,RVs,TNet).
```

Figure 5: Fragment of $S_5$, an SLP defining a prior over Bayesian nets (for a fixed set of random variables)

### 8.1 Imaginary Models

Having carefully filtered out unwanted models, we find that it is convenient to re-admit them to the model space when we implement our posterior sampling algorithm. However all these *imaginary models*, which previously did not have a probability defined for them, will now get probability zero. Doing this ensures that generating a new proposed model using $q(M^i, M^*)$ is simple. If the proposed model $M^*$ is imaginary then we will never accept it: since $p(M^*) = 0$ we have $\alpha(M^i, M^*) = 0$. An analogous approach exists in analysis, where it is often easier to do real analysis within the space of complex numbers.

Recall that the distribution $p_{(\lambda, S, \leftarrow model(M))}$ over atoms $model(m)$ is generated by marginalisation from a distribution of the same name over refutations of $\leftarrow model(M)$. Extending our definition to include zero probability imaginary models amounts to extending this underlying distribution on refutations to also include zero probability SLD-derivations that end in `fail`. It turns out that this last distribution on derivations is the most convenient to work with. Note that each derivation corresponds to a leaf in an SLD-tree. We will associate a leaf corresponding to a failure derivation with `fail` and a leaf corresponding to a refutation with the model yielded by that refutation. Non-leaf nodes of the SLD-tree will be associated with goals (see Fig 6).

### 8.2 The Transition Kernel

We can generate a new derivation (yielding a new proposed model $M^*$) from the derivation which yielded the current model $M^i$ as follows.

1. Backtrack one step to the most recent choice point in the SLD-tree

2. We then probabilistically backtrack as follows: If at the top of the tree stop. Otherwise backtrack one more step to the next choice point with probability $p$.

3. Use loglinear sampling to generate a derivation from the choice point chosen in previous backtracking step. However, in the first step of sampling we may not choose the branch that leads back to $M^i$.

If the derivation so found ends in `fail` then $M^*$ is an imaginary model, so $p(M^i) = 0$, $\alpha(M^i, M^*) = 0$ and we stay at $M^i$. The parameter $p$ controls the size of steps; if $p = 1$, we always restart from the top of the tree.

Now, we must calculate $\alpha(M^i, M^*)$ when $M^*$ turns out to be a real model. First let $G$ be the deepest common parent goal for $M^i$ and $M^*$. (Strictly, we should say "for the derivations that yield $M^i$ and $M^*$", but from now on we will abbreviate in this way.) There is only one way we can get from $M^i$ and $M^*$: backtracking to $G$ and then reaching $M^*$ from $G$. We can not go via some parent of $G$ (such as $G_0$ in Fig 6) since we have prohibited going down the tree the same way we backtracked up it.

Suppose that $M^i$ is at depth $N^i$ in the SLD-tree, then the probability of backtracking through $n^i$ choice points is $p^{n^i-1}(1-p)$ for $1 \leq n^i < N^i$ and $p^{N^i-1}$ for $n^i = N^i$. Let $B^i$ be the random variable that gives the number of backtracks from $M^i$. Similarly, we have $B^*$ for $M^*$. Then we find that $P(B^* = n^*)/P(B^i = n^i) = p^{(n^*-n^i)}$ whether $G$ is at the top-level choice point or not.

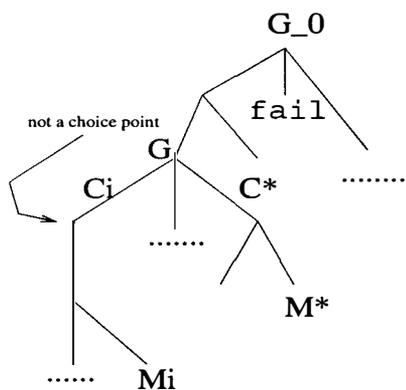

Figure 6: Jumping from $M^i$ to $M^*$ in the SLD-tree

The probability of reaching $M^*$ starting from $G$ is $\psi_{(\lambda,S,G)}(M^*)/(1 - l(C^i))$ where $C^i$ is the clause which is used at $G$ to get to $M^i$. So $q(M^i, M^*) = P(B^i = n^i)\psi_{(\lambda,S,G)}(M^*)[1 - l(C^i)]^{-1}$. Swapping $i$ and $*$ symbols gives us $q(M^*, M^i)$. The SLD-tree in Fig 6 shows an example, where $n^i = n^* = 2$. Note the imaginary model under the top-level goal. Next, note that $\psi_{(\lambda,S,G_0)}(M)/\psi_{(\lambda,S,G)}(M)$ is constant for any $M$ below $G$, since cancelling out the labels of clauses that get us from $G$ to $M$ in $\psi_{(\lambda,S,G_0)}(M)$ just leaves us with those that get us as far as $G$. In particular

$$\frac{\psi_{(\lambda,S,G_0)}(M^*)}{\psi_{(\lambda,S,G)}(M^*)} = \frac{\psi_{(\lambda,S,G_0)}(M^i)}{\psi_{(\lambda,S,G)}(M^i)}$$

Finally note that $p_{(\lambda,S,G_0)}(M^*)/p_{(\lambda,S,G_0)}(M^i) = \psi_{(\lambda,S,G_0)}(M^*)/\psi_{(\lambda,S,G_0)}(M^i)$, since we are actually dealing with derivations that yield models, and the normalising $Z$ factor cancels out. Putting all this together:

$$\frac{q(M^*, M^i)}{q(M^i, M^*)} \frac{p_{(\lambda,S,G_0)}(M^*)}{p_{(\lambda,S,G_0)}(M^i)}$$
$$= \frac{q(M^*, M^i)}{q(M^i, M^*)} \frac{\psi_{(\lambda,S,G_0)}(M^*)}{\psi_{(\lambda,S,G_0)}(M^i)}$$
$$= \frac{P(B^* = n^*)\psi_{(\lambda,S,G)}(M^i)[1 - l(C^*)]^{-1}}{P(B^i = n^i)\psi_{(\lambda,S,G)}(M^*)[1 - l(C^i)]^{-1}} \frac{\psi_{(\lambda,S,G_0)}(M^*)}{\psi_{(\lambda,S,G_0)}(M^i)}$$
$$= p^{(n^*-n^i)} \frac{1 - l(C^i)}{1 - l(C^*)}$$

So for real $M^*$, we have

$$\alpha(M^i, M^*) = \min\left\{p^{(n^*-n^i)} \frac{1 - l(C^i)}{1 - l(C^*)} \frac{\ell(M^*)}{\ell(M^i)}, 1\right\} \quad (1)$$

where $\ell(M)$ is the likelihood of the model: the probability of the observed data given the model, which we assume is at least defined (if not quickly calculable). Note that we always jump if the following three conditions are met (i) $\ell(M^*) \geq \ell(M^i)$ ($M^*$ fits the data at least as well as $M^i$), (ii) $n^* \geq n^i$ ($M^*$ is at least as deep as $M^i$) and (iii) $l(C^*) \geq l(C^i)$.

We can propose models quickly since we allow imaginary models. We can also compute the acceptance probability (1) easily. The main deficiency is that we may move very slowly through the space of real models if our SLP is highly constrained, leading to slow convergence.

## 9 Implementation and Experiments

Let us briefly show how SLPs can be implemented in Prolog, just to see how easy it is. The SLP fragment in Fig 4 is translated to the Prolog code in Fig 7. Each probabilistic predicate gets 3 extra arguments: one is a clause label and the other two (which are hidden in Fig 7 by DCG notation), are accumulators which pass around a list of the clause labels used and the potential of the derivation. This ugly but efficient implementation can easily by generated by a source-to-source transformation from more aesthetically pleasing code.

```
theory(main,In,Out) -->
 select_clause(theory_2,ClauseNum),
 theory(ClauseNum,In,Out).
theory(theory_2_1,Done,NicelyDone) -->
 {\+ Done = []},
 {nice_all(Done,NicelyDone)}.
theory(theory_2_2,RulesSoFar,Done) -->
 rule(Rule),
 {newrule(RulesSoFar,Rule)},!,
 theory(main,[Rule|RulesSoFar],Done).

labels(theory_2,[cdp(theory_2_1,0.1,0.1) ,
  cdp(theory_2_2,0.9,1)]).
```

Figure 7: An SLP in Prolog

We have yet to do real experiments to test whether our sampling algorithm converges on the true posterior, but at least have a working implementation that is reasonably efficient, and which can be used to explore the consequences of altering various parameters. Running the algorithm with no data (so the likelihoods do not need calculating) using the $S_4$ prior over logic programs took a little over 9 seconds of CPU time to produce (and write out) 10000 samples on a Pentium 233MHz running Yap Prolog. This involved 546 acceptances of a proposed $M^*$, an acceptance rate of only 5.46% and involved 337 distinct logic programs. When run using a data set of 5 positive and 5 negative examples and a simple 10% classification noise likelihood function, 10000 samples took 11.5 seconds, involving 451 jumps and 465 distinct logic programs. These runs were done using a backtrack probability of $p = 0.8$. Reducing $p$ to 0.3 produced only 39 jumps out of 10000.

## 10  Conclusions and Future Directions

We have defined a number of algorithms for SLPs, together with relevant mathematical analysis. This goes some way to establishing that SLPs can be a useful framework for representing and reasoning with complex probability distributions. We view the application to Bayesian machine learning as being the most promising area for future research. The definition of a general-purpose and practical transition kernel is probably the paper's major contribution. However, it remains to be proven by rigorous experimentation that our posterior sampler produces better results than more conventional search-based approaches. Also, we have also yet to give a proper account of termination when sampling from SLPs.

In this paper, we have only considered priors over structures, not parameters; but it is easy to embed built-in predicates in the Prolog code to generate e.g. samples from a Dirichlet. More interestingly, there is the possibility of combining the likelihood with the prior to generate a posterior *in the same form as the prior*. It is easy to construct an *impractical* SLP for the posterior of the form:

```
posterior(Model) :-
   prior(Model), likelihood(Model).
```

The interesting question is whether this impractical definition can be transformed into a usable representation. One problem here, is that the size of an efficient representation of the posterior is likely to explode, given that the posterior is generally more complex than the manually-derived prior.

We conclude by pointing to (Cussens, 2000) where a much more detailed account of SLPs is given, and where the EM algorithm is applied to estimate SLP parameters from data.

### Acknowledgments

Special thanks to Blaise Egan for preventing me from reinventing the wheel and Gillian Higgins for putting up with me. Thanks also to Stephen Muggleton and Suresh Manandhar for useful discussions.